\documentclass[10pt,twocolumn,letterpaper]{article}

\usepackage{iccv}
\usepackage{authblk}
\usepackage{times}
\usepackage{epsfig}
\usepackage{graphicx}
\usepackage{amsmath}
\usepackage{amssymb}
\usepackage{enumitem}
\usepackage{multirow}
\usepackage{booktabs}
\usepackage{bbm}
\usepackage[T1]{fontenc}
\usepackage[utf8]{inputenc}
\usepackage[english]{babel}
\usepackage[autostyle, english=american]{csquotes}
\usepackage{subcaption}
\usepackage{amsmath}
\usepackage{xcolor}
\usepackage{makecell}
\usepackage{multirow}
\usepackage{booktabs}
\usepackage{amssymb}
\MakeOuterQuote{"}
\usepackage{bbm}

\usepackage[breaklinks=true,bookmarks=false]{hyperref}

\iccvfinalcopy 

\def\iccvPaperID{****} 

\ificcvfinal\fi
\ificcvfinal\fi
\begin{document}

\title{Solution for Point Tracking Task of ECCV 2nd Perception Test Challenge 2024}

\author{
Yuxuan Zhang\textsuperscript{1, 2},
Pengsong Niu\textsuperscript{1},
Kun Yu\textsuperscript{1},
Qingguo Chen\textsuperscript{2},
Yang Yang\textsuperscript{1, $\thanks{Corresponding Author}$} 
}

\affil{
 $^1$Nanjing University of Science and Technology
 $^2$Alibaba International Digital Commerce Group
 
}
\setlength{\intextsep}{2pt}
\setlength{\abovecaptionskip}{2pt}
\maketitle

\begin{abstract}
This report introduces an improved method for the Tracking Any Point~(TAP), focusing on monitoring physical surfaces in video footage. Despite their success with short-sequence scenarios, TAP methods still face performance degradation and resource overhead in long-sequence situations. To address these issues, we propose a simple yet effective approach called Fine-grained Point Discrimination~(\textbf{FPD}), which focuses on perceiving and rectifying point tracking at multiple granularities in zero-shot manner, especially for static points in the videos shot by a static camera. The proposed FPD contains two key components: $(1)$ Multi-granularity point perception, which can detect static sequences in video and points. $(2)$ Dynamic trajectory correction, which replaces point trajectories based on the type of tracked point. Our approach achieved the second highest score in the final test with a score of $0.4720$.
\end{abstract}

\section{Introduction}
Deep learning has significantly advanced computer vision tasks, particularly in visual understanding~\cite{MeinhardtKLF22, YangFZLJ21, YangWZX019, YangWZL018, YangYBZZGXY23, 0074ZGGZ22, YangZSX23, YangZZX019, YangZZXJY23}. Frame-by-frame analysis of each pixel in video streams and tracking their motion trajectories is a fundamental task in computer vision~\cite{HuangZHSZ22, LiZHHGC23}, playing a crucial role in various applications such as video object segmentation, tracking, and understanding the physical world. Traditionally, most methods rely on optical flow data to solve the correspondence problem between two consecutive frames and address occlusion, while often overlooking long-range temporal information~\cite{NeoralSM24, YangHGXX23}. To expand the range of tracking points and further track various objects in the video, several attempts~\cite{DoerschYVG0ACZ23_2023, lemoing2024dense_2024, Li_2024} have proposed allowing users to specify arbitrary points throughout the video, formalizing the task as tracking any point. 

Different from the methods based on optical flow, the primary challenge of TAP is modeling each tracked point over time, which may involve unforeseen occlusion~\cite{lemoing2024dense_2024}. Mainstream methods typically employ a sliding window approach, allowing points in different frames within the same window to share information across the temporal axis. TAPIR~\cite{DoerschYVG0ACZ23_2023} implements a two-stage process of matching and refinement to independently query and predict detailed point trajectories and features using local information. However, these attempts fail to consider the correlations and physical relationships between the tracked points, resulting in subpar performance in challenging point tracking. DOT~\cite{lemoing2024dense_2024} aims to integrate optical flow and point tracking techniques to expedite the attainment of accurate point trajectories. TAPTR~\cite{Li_2024} addresses the problem by modeling each point as a query, allowing for natural layer-by-layer updates of its coordinates and visibility through self-attention along the temporal dimension. TAPTR clarifies the information in the TAP task and enhances interactions between query points through self-attention operations along the temporal dimension to improve performance.  However, these methods still face performance degradation and resource overhead in long-sequence situations.

\begin{figure}[h]
    \centering
    \includegraphics[width=82mm]{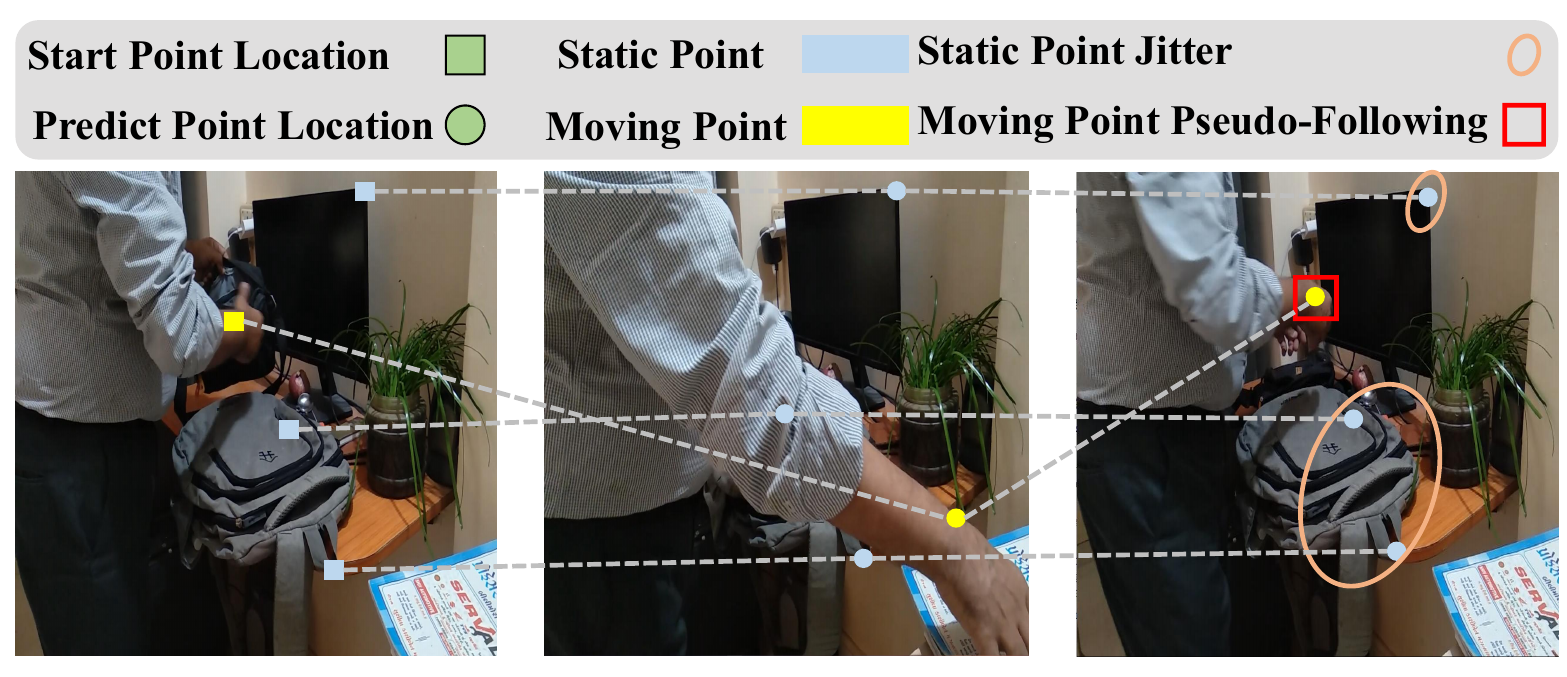}\\
    \caption{The common issues of mainstream TAP methods.}
 \label{fig:PT_issues}
\end{figure}

We aim to utilize off-the-shelf models to achieve improved tracking trajectories in a zero-shot manner. 
By comparing various classical methods for predicting point trajectories in the same video, we identified two common issues, as illustrated in Figure 1~\ref{fig:PT_issues}: \textbf{i) Static Point Jitter}: Points occasionally exhibited minor positional jitter, even when using state-of-the-art~(SOTA) methods that account for the correlations of query points.\textbf{ii) Moving Point Pseudo-Following}: Due to the length limitations of existing methods' sliding windows, predictions for most moving points often exhibit drifting. 

\begin{figure*}[h]
  \centering
  \includegraphics[trim=0 0 0 0, clip, width=0.90\textwidth]{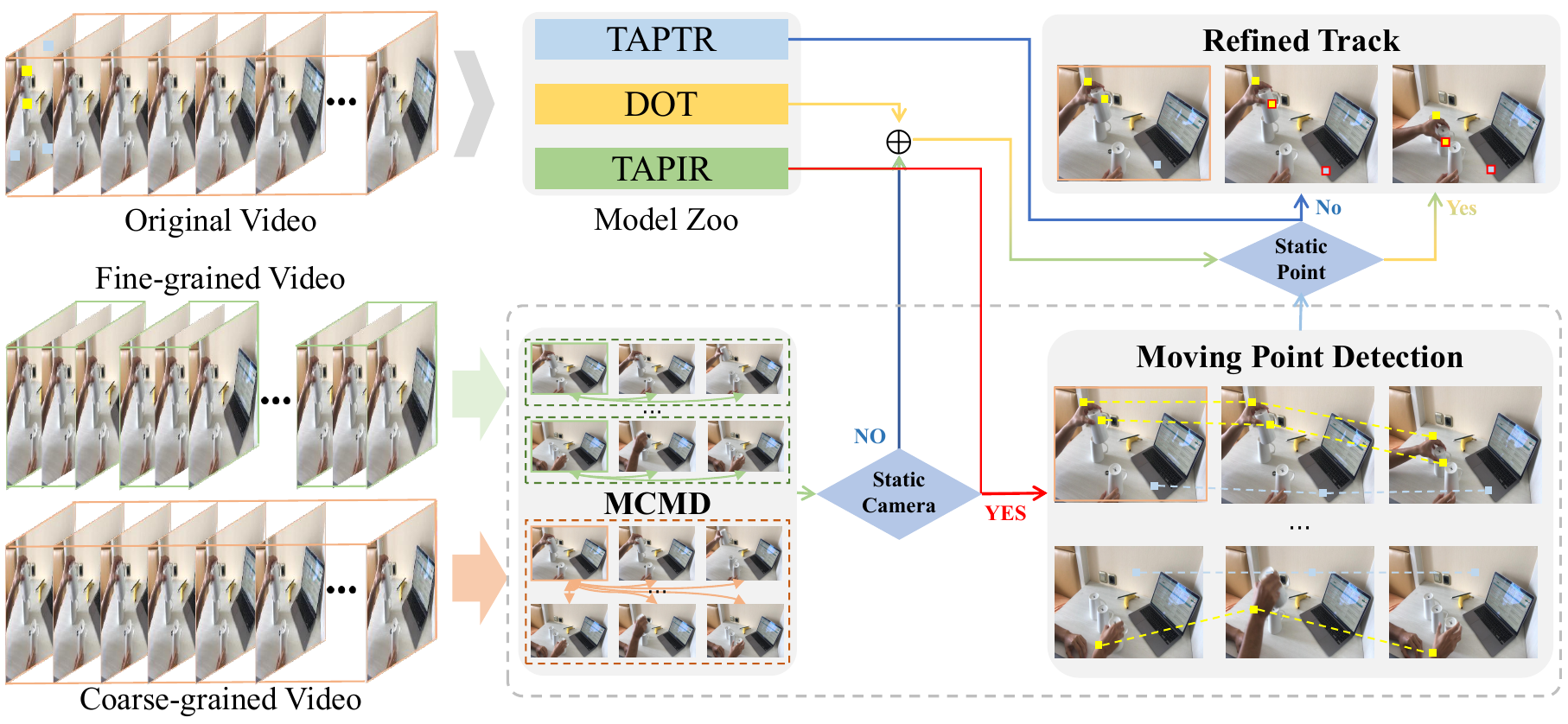}
  \caption{\textbf{Framework of the proposed Fine-grained Point Discrimination~(\textbf{FPD}).} MCMD: multi-granularity camera motion detection. We first initialize the tracks of each point using three SOTA TAP methods~(i.e., TAPIR, DOT, and TAPTR). For the original video sequence, we conduct a hierarchical assessment at both the camera and point levels. At the camera level, we use a multi-granularity motion detection algorithm to determine if the video is stationary. At the point level, we apply a Moving Point Detection algorithm to identify stationary predicted points. Based on these evaluations, we use dynamic trajectory correction to refine the tracks and achieve the final results.}
  \label{fig:overview}
\end{figure*}

To address these issues, we propose an effective two-stage approach named \textbf{FPD}. The proposed FPD consists of two components: \textbf{i) Multi-granularity Point Perception}~(MPP), which employs a multi-granularity camera motion detection~(MCMD) algorithm and a moving point detection~(MPD) algorithm to identify both static video and points. \textbf{ii) Dynamic Trajectory Correction}~(DTC), which replaces point trajectories based on the type of tracked point. Specifically, we take the original point trajectory prediction for moving camera video. For static camera videos, we rely on the discrimination results of stationary points to replace point trajectories. To this end, the contributions can be summarized as follows:

\begin{itemize}
\item We have proposed a simple yet effective zero-shot TAP method, which leverages the off-the-shelf models and can further distinguish between moving and static points to obtain optimal tracks.
\item Extensive experiments demonstrate the superiority of the proposed FPD, which achieved 2nd place with a score of 0.4720 in the final test.
\end{itemize}

\section{Method}
\subsection{Preliminary}
Given a video sequence and a set of points to track, the TAP task aims to determine whether each point is visible and, if so, to identify its location in each frame. Specifically, give an video sequence $\mathcal{V}$, we define $\mathcal{V} = \{v_t\}_{t=1}^T$, where $T$ denotes the number of frames. The query points set can be defined as $\mathbf{Q}= \{ q_{e}^{i}\}_{i=1}^{M} = \{ (x_{e}^{i}, y_{e}^{i})\}_{i=1}^{M} $, where $M$ is the number of query points and $e$ is the index indicating the time stamp when the tracking point first emerges or starts to be tracked. The purpose of TAP is to track the point across the video to obtain its trajectory and visibility sequence, which the following equation can describe:
\begin{align} \label{eq:purpose}
&\mathcal{L}^{i} = \{q_{t}^{i}\}_{t=1}^{T} = \{ (x_{t}^{i}, y_{t}^{i})\}_{t=1}^{T},\\
&\text{Vis} = \{\text{vis}_{t}^{i}\}_{t=1}^{T}, \text{vis} \in \{0, 1\}.\nonumber
\end{align}

\subsection{Base Model}
In this work, we employ three models~(i.e., TAPIR+~\cite{Hongpeng_2024}, DOT, TAPTR) to achieve optimal results. TAPIR+ builds on TAPIR by incorporating Multi-granularity Camera Motion Detection and CMR-based point trajectory prediction for camera-level trajectory refinement. Meanwhile, DOT extracts tracks from key motion boundaries and computes initial estimates of a dense flow field and visibility mask via nearest-neighbor interpolation. It then refines these estimates with a learnable optical flow estimator that handles occlusions and is trainable on synthetic data with ground-truth correspondences, further optimizing TAPIR+'s CMR-based point trajectory prediction. TAPTR clarifies the information in the TAP task and enhances interactions between query points through self-attention operations along the temporal dimension to improve performance. However, these methods each have drawbacks regarding stationary points and moving points in static camera, resulting in suboptimal outcomes. 

\subsection{The Proposed FPD}
To solve the static point jitter and moving point pseudo-following, we focus on perceiving and rectifying point tracking at multiple granularities. The proposed method consists of two primary components, as illustrated in Figure~\ref{fig:overview}, which will be discussed in the following sections. 

\subsubsection{Multi-granularity Point Perception}
To effectively perceive the trajectories of query points across different videos, we conduct multi-level perception at both the camera and point levels and propose multi-granularity point perception. Specifically, the proposed multi-granularity point perception contains multi-granularity camera motion detection~(MCMD) and moving point detection~(MPD). 

The purpose of the MCMD is to distinguish between static and moving camera shots in various videos. This algorithm concurrently considers both video-level~(coarse-grained) and clip-level segment~(fine-grained) analyses, where the original video $\mathcal{V}$ is partitioned into multiple 5-second clip segments $\mathcal{V}=\{v^1, v^2, ..., v^n\}$, $n$ represents the number of clips the video has been divided into. Each frame in videos and clips undergoes a grayscale transformation. To determine whether a video clip is a motion shot, we designate the first frame as the reference frame and compute the Structural Similarity Index Measure~(SSIM)~\cite{SSIM_2020} between subsequent frames and the reference frame. These scores are calculated using functions from the \textit{skimage} library, with lower SSIM scores indicating greater dissimilarity between frames. The process of calculating
the moving score can be represented by:
\begin{equation}
    \textbf{ms}_\mathcal{V} =\mathbbm{1}\left(\left(\frac{1}{T}\sum_{t}^{T} \mathbbm{1}(SSIM(\textbf{v}_1,\textbf{v}_t)<\lambda)\right) > \eta \right),
\end{equation}
where $\textbf{v}_t$ denotes $t$-th frame in video $\mathcal{V}$, and $\mathbbm{1}$ represents the indicator function. $\textbf{y}_\textbf{v}$ indicates the initial results of MCMD, with $0$ representing static and $1$ representing moving. We consider both fine-grained and course-grained video clips and the final moving score can be  formalized as follows:
\begin{align}
    &\textbf{ms}_\mathcal{V}^{fg} = \textbf{ms}_{\textbf{v}^1}^{fg} \vee \textbf{ms}_{\textbf{v}^2}^{fg} \vee \ldots \vee \textbf{ms}_{\textbf{v}^n}^{fg},\\
    &\textbf{ms}_\mathcal{V} = \textbf{ms}_\mathcal{V}^{cg} \land \textbf{ms}_\mathcal{V}^{fg},\nonumber
\end{align}
where $\textbf{ms}_\mathcal{V}^{fg}$, $\textbf{ms}_\mathcal{V}^{cg}$ means the fine-grained and course-grained video moving score.

The purpose of the MPD is to leverage the predicted tracks to further distinguish between moving and static points. Given the track $\mathcal{L}^{i}$ in Eq. (\ref{eq:purpose}) predicted by the base model, we calculate the variance of the coordinates of all visible frames to assess whether the current point is stationary.
The entire process can be formalized as follows:
\begin{align}
    &\sigma_{\mathcal{S}}^i = \sqrt{\frac{1}{n} \sum_{i=1}^{T} (\mathcal{S}_i - \mu)^2},\\
    &\textbf{d}^i = \mathbbm{1} (\sigma_{\textbf{x}}^i  \land \sigma_{\textbf{y}}^i) <\rho,\nonumber
\end{align}
where ${S}_i$ denotes the set of coordinates for the predicted tracks. $\textbf{x}$ and $\textbf{y}$ represents the coordinates in $\mathcal{L}^{i}$. $\rho$ represents the deviation of the coordinates. 

\subsubsection{Dynamic Trajectory Correction}
Based on the aforementioned camera and point-level perception, we integrate the predictions from different base models according to the perceived results. We define the tracks of $i$-th point predicted by the three models as follows:
\begin{align}
    & \mathcal{L}^{i}_{TAPIR+} = \{q_{t}^{i}\}_{t=1}^{T},\nonumber\\
    & \mathcal{L}^{i}_{DOT} = \{q_{t}^{i}\}_{t=1}^{T},\\
    & \mathcal{L}^{i}_{TAPTR} = \{q_{t}^{i}\}_{t=1}^{T}.\nonumber
\end{align}

For all points in moving camera videos, we use the tracks predicted by TAPTR and Dot. For static points in the static camera, we replace them with a mixture of tracks from TAPIR+, while for moving points, we use tracks from TAPTR for replacement.

\section{Experiments}
\textbf{Dataset.} As a zero-shot approach, our method relies on several pre-trained models~(i.e., TAPIR, DOT, TAPTR) and weights, with no additional data used for training. Concerning the pre-trained data, All the models adjust camera angles in the generation script of the public MOVi-E dataset to create a customized training dataset named MOVi-F. Both the validation and test sets use the official data provided.

\textbf{Metric.} The evaluation metric for this Task is the Average Jaccard (AJ), proposed in TAP-Vid~\cite{DoerschGMRSACZY22}.

\textbf{Implementation Detail.} All models are assessed using videos with a maximum resolution of $256 \times 256$, and subsequently, all TAP-Vid metrics are calculated within the same resolution of $256 \times 256$. Our approach is founded on zero-shot learning principles and uses the pre-trained model made available from the official code. Regarding hyperparameter settings, we respectively configure the similarity threshold as $\lambda_1=0.5$ and $\lambda_2=0.46$, where $\lambda_1$ and $\lambda_2$ represent the hyperparameters for coarse and fine granularity video segmentation, respectively.. And establish a deviation of the coordinates of $\rho=0.00125$.

\begin{table}[htp]
    \small
    \centering
    \caption{ The Average Jaccard of TAP on competition dataset. *: The experimental data is sourced from official authorities. sc\_sp: Static point from static camera video. sc\_mp: Moving point from moving camera video. mc\_mp: Moving point from moving camera video. }
    \begin{tabular}{c|ccccc}
    \toprule
    Method & sc\_sp AJ & sc\_mp AJ & mc\_mp AJ&  AJ \\
    \hline
    Baseline* & -  & -  & -  & 42.00  \\
    TAPIR & - & -  & - & 43.24  \\
    TAPIR+ & 43.01 & 21.53 & 31.86 & 45.78  \\
    TAPTR & 39.23 & \textbf{28.82} & 26.87 & 43.14  \\
    TAPIR+DOT & 42.83 & 27.35 & \textbf{32.35} & 46.12  \\
    \hline
    FPD & \textbf{43.90} & 24.83 & 32.13 & \textbf{47.20}  \\
    \toprule
\end{tabular}

\label{tab: compare}
\end{table}

\textbf{Comparison Methods Result.} Table \ref{tab: compare} shows the AJ score performance, from which we can observe that: $1)$ Compared to TAPIR+, both TAPTR and TAPIR+DOT demonstrate better performance on moving points in static videos, while their performance on stationary points in static videos declines. $2)$ The proposed FPD combines the strengths of existing methods and outperforms all TAP approaches. This phenomenon indicates that TAPIR+ offers improved performance in mitigating the jitter and drift of the static points.

\begin{table}[htp]
    \small
    \centering
    \caption{Ablation experiment.} 
    \begin{tabular}{c|ccccc}
    \toprule
    Method & sc\_sp AJ & sc\_mp AJ & mc\_mp AJ&  AJ \\
    \hline
    TAPIR+ & 43.01 & 21.53 & 31.86 & 45.78  \\
    + MPP  &43.23 &26.09 &32.05 & 46.70\\
    + DTC & 43.90 & 24.83 & 32.13 & 47.20  \\
    \toprule
\end{tabular}

\label{tab: abl}
\end{table}

\textbf{Ablation Study.} To evaluate the impact of each component in FPD, we perform additional ablation studies on the competition's test set. Table \ref{tab: abl} demonstrates the Average Jaccard score after adding components. The results indicate that the MPP significantly enhances the AJ value of static shot motion points, while the DTC further improves the overall AJ. However, the AJ score for moving points in static camera setups shows a slight decline, which may be due to the MPD discrimination criteria lacking precision and requiring further investigation.

\begin{table}[htp]
    \small
    \centering
    \caption{Parameter sensitivity analysis.}
    \begin{tabular}{c|ccccc}
    \toprule
     $\rho$ & sc\_sp AJ & sc\_mp AJ & mc\_mp AJ&  AJ \\
    \hline
    0.00100 & 42.89 & 23.53 & 31.46 & 46.19  \\
    0.00125  &43.23 &26.09 &32.05 & 46.70\\
    0.00150 & 37.80 & 24.05 & 32.13 & 47.12  \\
    0.00200 & 37.82 & 23.92 & 32.12 & 47.04  \\
    
    \toprule
    \end{tabular}

\label{tab:para}
\end{table}

\textbf{Parameter Sensitivity Analysis.} To verify the sensitivity of parameters, we conduct more experiments by tuning the crucial dissimilar frame percentage parameter $\rho$. Table \ref{tab:para} shows the AJ score on the test set of the competition. FDP achieves the best AJ score with $\rho$ = 0.00125, which indicates that FPD can effectively distinguish between moving points and static camera points, handle the jitter and drift, and improve the overall AJ score.

\section{Conclusion}
This report outlines our solution for the Point Tracking Task in the 2nd Perception Test Challenge at ECCV 2024. We proposed a zero-shot TAP approach, named FPD, which incorporates multi-granularity point perception and dynamic trajectory correction to enable fine-grained trajectory point detection, thereby effectively enhancing the overall performance of single-point object tracking.

{\small
\bibliographystyle{ieee_fullname}
\bibliography{PT}
}
\end{document}